\documentclass[graybox]{svmult}
\usepackage{mathptmx}
\usepackage{helvet}
\usepackage{courier}
\usepackage{type1cm}
\usepackage{makeidx}
\usepackage{graphicx}
\usepackage{multicol}
\usepackage{floatrow}
\usepackage[normalem]{ulem}
\usepackage[font={small}]{caption}
\usepackage[bottom]{footmisc}
\makeindex
\usepackage{multicol}
\usepackage{enumitem}
\usepackage[mathcal]{euscript}
\usepackage{subcaption}
\usepackage{siunitx}
\usepackage{mathtools}

\usepackage{array}
\newcolumntype{P}[1]{>{\centering\arraybackslash}p{#1}}
\newcolumntype{M}[1]{>{\centering\arraybackslash}m{#1}}
\usepackage[normalem]{ulem}
\newcommand{\tmax}{\ensuremath{t_{\text{max}}}\xspace}
\newcommand{\dmin}{\ensuremath{d_{\text{min}}}\xspace}
\usepackage{xspace}
\newcommand{\mPerSec}[1]{\SI[per-mode=symbol]{#1}{\meter\per\second}}
\newcommand{\degPerSec}[1]{\SI[per-mode=symbol]{#1}{\degree\per\second}}

\newcommand\moduleName[1]{\textsf{#1}\xspace}
\newcommand{\GP}{\moduleName{GP}}
\newcommand{\SM}{\moduleName{SM}}
\newcommand{\PP}{\moduleName{PP}}
\newcommand{\GA}{\moduleName{GA}}

\newcommand\blfootnote[1]{%
  \begingroup
  \renewcommand\thefootnote{}\footnote{#1}%
  \addtocounter{footnote}{-1}%
  \endgroup
}

\usepackage{hyperref}
\hypersetup
{
    colorlinks = true,
    citecolor = blue,
    linkcolor = blue,
}

\graphicspath{ {./images/} }

\usepackage{booktabs}
\newcommand{\ra}[1]{\renewcommand{\arraystretch}{#1}}
\usepackage{seqsplit}
\usepackage{multirow, makecell}
\usepackage{amsmath}
\usepackage{amssymb}
\usepackage{amsfonts}
\usepackage{bbm}

\usepackage{algorithm}
\usepackage[noend]{algpseudocode}
\usepackage{algorithmicx}

\usepackage{wrapfig}
\usepackage[dvipsnames]{xcolor}
\usepackage{color}

\newcommand{\Pmap}{\ensuremath{\mathcal{M}^{\text{P}}}\xspace}
\newcommand{\Omap}{\ensuremath{\mathcal{M}^{\text{O}}}\xspace}
\newcommand{\Nmap}{\ensuremath{\mathcal{M}^{\text{NC}}}\xspace}

\usepackage{titlesec}
\titlespacing*{\section}{0pt}{0.6\baselineskip}{0.6\baselineskip}
\titlespacing*{\subsection}{0pt}{0.6\baselineskip}{0.6\baselineskip}
\titlespacing*{\subsubsection}{0pt}{0.6\baselineskip}{0.6\baselineskip}
\usepackage{chngcntr}
\counterwithout{equation}{chapter}
\usepackage{multirow, makecell}

\begin{document}

\title*{A Planning Framework for Persistent, Multi-UAV Coverage with Global Deconfliction}
\titlerunning{A Planning Framework for Persistent, Multi-UAV Coverage}

\author{
Tushar Kusnur*,
Shohin Mukherjee*,
Dhruv Mauria Saxena,
Tomoya Fukami,
Takayuki Koyama,
Oren Salzman,
and Maxim Likhachev \\\vspace*{5pt}
*equal contribution
}
\authorrunning{ } 
\institute{T. Kusnur
\and S. Mukherjee
\and D. M. Saxena
\and O. Salzman
\and M. Likhachev
\at Carnegie Mellon University, Pittsburgh, PA, USA.
\email{{tkusnur, shohinm, dsaxena, osalzman, mlikhach}@andrew.cmu.edu}
\and T. Fukami
\and T. Koyama
\at Mitsubishi Heavy Industries, Ltd., Tokyo, Japan.
\email{{tomoya_fukami, takayuki_koyama}@mhi.co.jp}}
%
\maketitle
\vspace{-60pt}
\abstract*
{
Planning for multi-robot coverage seeks to determine collision-free paths for a fleet of robots, enabling them to collectively observe points of interest in an environment.
Persistent coverage is a variant of traditional coverage where coverage-levels in the environment decay over time. Thus, robots have to continuously revisit parts of the environment to maintain a desired coverage-level.
Facilitating this in the real world demands we tackle numerous subproblems.
While there exist standard solutions to these subproblems, there is no complete framework that addresses all of their individual challenges as a whole in a practical setting.
We adapt and combine these solutions to present a planning framework for persistent coverage with multiple unmanned aerial vehicles (UAVs).
Specifically, we run a continuous loop of goal assignment and globally deconflicting, kinodynamic path planning for multiple UAVs.
We evaluate our framework in simulation as well as the real world. In particular, we demonstrate that (i) our framework exhibits graceful coverage---given sufficient resources, we maintain persistent coverage; if resources are insufficient (e.g., having too few UAVs for a given size of the enviornment), coverage-levels decay slowly and (ii) planning with global deconfliction in our framework incurs a negligibly higher price compared to other weaker, more local collision-checking schemes.  (Video: \url{https://youtu.be/aqDs6Wymp5Q})
}
\abstract
{
Planning for multi-robot coverage seeks to determine collision-free paths for a fleet of robots, enabling them to collectively observe points of interest in an environment.
Persistent coverage is a variant of traditional coverage where coverage-levels in the environment decay over time. Thus, robots have to continuously revisit parts of the environment to maintain a desired coverage-level.
Facilitating this in the real world demands we tackle numerous subproblems.
While there exist standard solutions to these subproblems, there is no complete framework that addresses all of their individual challenges as a whole in a practical setting.
We adapt and combine these solutions to present a planning framework for persistent coverage with multiple unmanned aerial vehicles (UAVs).
Specifically, we run a continuous loop of goal assignment and globally deconflicting, kinodynamic path planning for multiple UAVs.
We evaluate our framework in simulation as well as the real world. In particular, we demonstrate that (i) our framework exhibits graceful coverage---given sufficient resources, we maintain persistent coverage; if resources are insufficient (e.g., having too few UAVs for a given size of the enviornment), coverage-levels decay slowly and (ii) planning with global deconfliction in our framework incurs a negligibly higher price compared to other weaker, more local collision-checking schemes.  (Video: \url{https://youtu.be/aqDs6Wymp5Q})
}
\section{Introduction}
\label{sec:1}
Traditional robot-coverage is the problem of determining a collision-free path for a robot that covers all points of interest in an environment~\cite{galceran2013survey}.
\emph{Persistent} coverage seeks to continuously maintain a desired \emph{coverage-level} over an environment~\cite{persistentFranco2013, persistentMellone2018,smith2011persistent}.
%
\blfootnote{\hspace{-3pt}\textbf{Acknowledgement}:
This work was sponsored by Mitsubishi Heavy Industries, Ltd.
The authors would also like to thank Daniel Feshbach for help with creating figures.}
%
In our case, coverage-levels degrade over time, thereby increasing the urgency with which points must be revisited.
For example, consider a floor-cleaning robot---once a part of the floor is cleaned, more dust will eventually collect over it and thereby decrease its coverage-level. 
There has been a rise in the use of robots to perform tasks cast as persistent-coverage problems, such as environmental monitoring, exploration, inspection, post-disaster assessment, monitoring traffic congestion over a city, etc.~\cite{nedjati2016complete, smith2011persistent, srinivasan2004airborne, teixeira2018autonomous}.

In general, coverage path-planning is closely related to the intractable \emph{covering-salesman problem}, where an agent must visit neighborhoods of multiple cities while minimizing travel-distance~\cite{arkin1994approximation}.
Extending this to multiple robots requires collision avoidance (with both static and dynamic obstacles), which becomes computationally expensive as the number of robots increase.
For persistent coverage, additional algorithmic challenges emerge since robots need to continuously revisit parts of the environment to maintain coverage-levels. For a single robot tasked with persistent coverage, there are broadly two decisions to be made:
\begin{enumerate}[label={\textbf{Q\arabic*}},leftmargin=1.0cm]
    \item \label{task:ga} Where should the robot go next?
    \item \label{task:gp} How should it get there?
\end{enumerate}
We refer to~\ref{task:ga} and~\ref{task:gp} as \emph{goal assignment} and \emph{goal planning} respectively.
Additional questions arise when dealing with multiple robots:
\begin{enumerate}[label={\textbf{Q\arabic*}},leftmargin=1.0cm]
    \setcounter{enumi}{2}
    \item \label{task:pp} How do we plan for multiple robots?
    \item \label{prop:deconf} How do we avoid inter-robot collisions?
\end{enumerate}
Assuming a centralized planner, we address \ref{task:pp} by sequential, decoupled planning instead of planning in the joint state-space of all robots for tractability.
Specifically, we use \emph{prioritized planning}~\cite{erdmann1987multiple}.

We use the notion of \emph{committed plans} to tie our answers to these questions together.
These are defined as parts of plans that robots commit to execute, in contrast to the rest of their plans which may be modified after reassessing the environment.
We maintain the invariant of \emph{global deconfliction}---no new committed plan intersects any other existing committed plan in space or time, thereby answering \ref{prop:deconf}.


\begin{figure}[t]
    \centering
    \begin{subfigure}{.61\textwidth}
        \includegraphics[width=\textwidth]{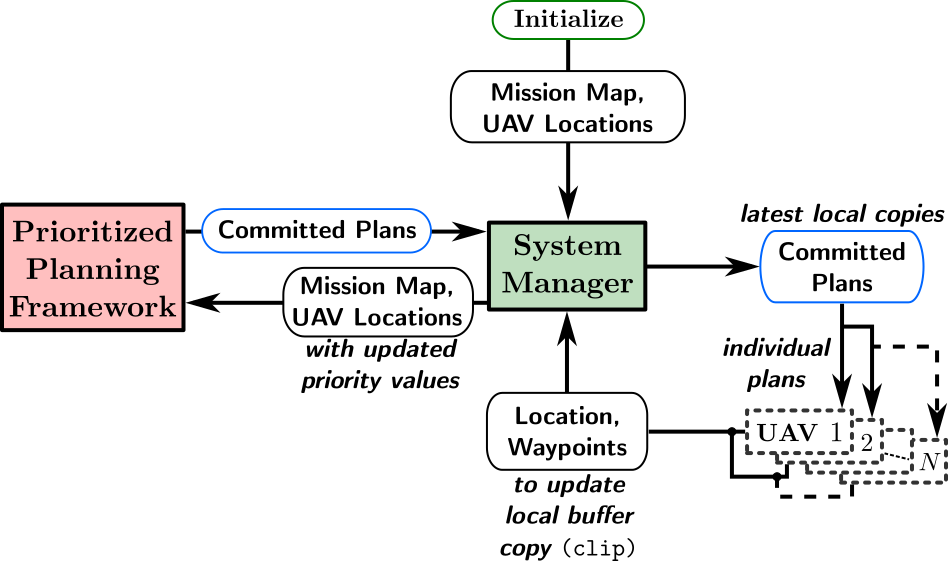}
        \caption{}
        \label{fig:sm_pp_uav}
    \end{subfigure}\hfill
    \begin{subfigure}{0.36\textwidth}
        \includegraphics[width=\textwidth]{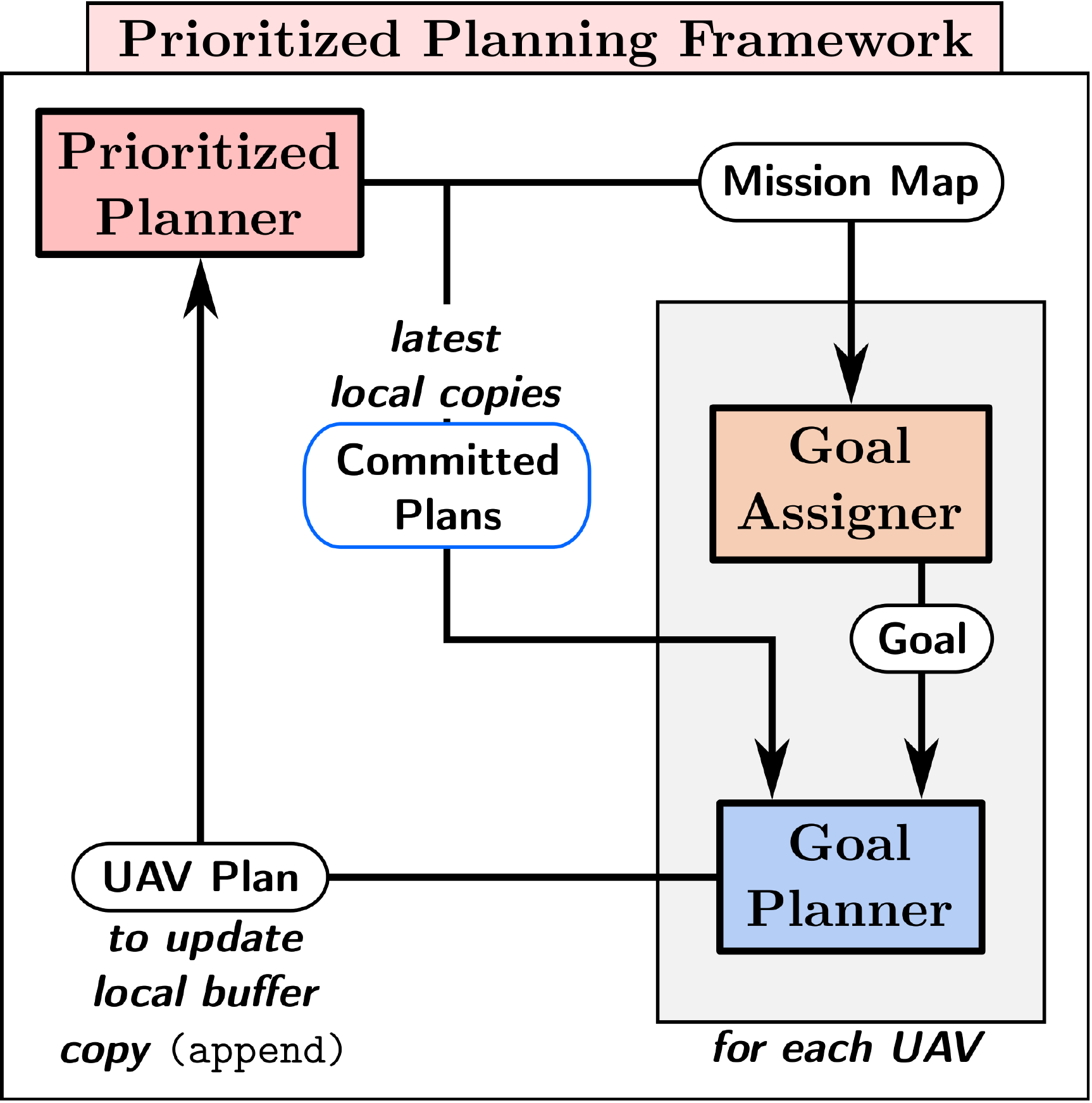}
        \caption{}
        \label{fig:pp_ga_gp}
    \end{subfigure}
    \caption{\small{(\subref{fig:sm_pp_uav}) The System Manager (\SM) communicates with all UAVs; it sends up-to-date copies of committed plans to corresponding UAVs and updates them using information received from the Prioritized Planner (\PP) (\subref{fig:pp_ga_gp}) Prioritized planning framework: For each UAV, the Goal Assigner (\GA) selects the next goal using the up-to-date map from the \PP and the Goal Planner (\GP) then plans a feasible path to this goal, which is appended to its committed plan}}
    \label{fig:block_diagram}
\end{figure}

In this paper, we present a planning framework to address all the above questions.
A block diagram of the complete framework is presented in Fig.~\ref{fig:block_diagram}.
To the best of our knowledge, this work is the first to answer all of these questions in unison for real-world, persistent coverage with multiple robots.
For our application, these robots are Unmanned Aerial Vehicles (UAVs).
The \emph{System Manager} (\SM, Sec.~\ref{subsec:SM}) is our communication interface between the centralized planner and individual UAVs.
It is responsible for back-and-forth communication of up-to-date information between the two.
The \emph{Prioritized Planner} (\PP, Sec.~\ref{subsec:PP}) is responsible for answering \ref{task:pp}.
Thereafter, the \emph{Goal Assigner} (\GA, Sec.~\ref{subsec:GA}) answers \ref{task:ga} and specifies the location the next UAV should fly to. 
This is fed to the \emph{Goal Planner} (\GP, Sec.~\ref{subsec:GP}), which plans a kinodynamically feasible path for the UAV to the goal, answering \ref{task:gp}.
Part of this plan is appended to the existing committed plan of the UAV.
The \PP, which operates continuously in a loop, then begins the next planning cycle, answering \ref{task:pp} again.

We only append a part of computed plans to committed plans of UAVs (further detailed in Sec.~\ref{sec:4}).
This is done so that we maintain committed plans of a pre-specified maximum duration (denoted by \tmax)\footnote{This duration depends on the type and capability of the robots, the number of robots, and the coverage map.}.
There is a trade-off involved in making \tmax too short or too long. Sec.~\ref{sec:4} provides more insight into how and why this arises.
We rely on globally deconflicting paths for UAVs since other, more local collision-checking mechanisms have a higher risk of causing \emph{deadlocks}\footnote{Two or more robots are in a \emph{deadlock} if they are not in collision, but executing any valid action for any one of the robots would cause them to collide with another. Hence, they remain stationary.}.
Some other approaches utilise conflict detection and resolution systems to repair conflicting paths~\cite{barnier2012trajectory}.
Our approach solves for deconflicting paths by construction and maintains the invariant of global deconfliction.

In Sec.~\ref{sec:2}, we contextualize our work among the Orienteering Problem (OP)~\cite{vansteenwegen2011orienteering} and the Vehicle Routing Problem (VRP)~\cite{toth2002vehicle}, coverage path-planning~\cite{galceran2013survey}, and frontier-based exploration~\cite{yamauchi1998frontier}.
Sec.~\ref{sec:3} formalizes the persistent-coverage problem we aim to solve and introduces the notion of priorities over coverage zones.
Sec.~\ref{sec:4} provides details about our planning framework and each of its constituent parts.
In Sec.~\ref{sec:5}, we present results of the performance of our framework in both simulated and real-world environments.
Finally, Sec.~\ref{sec:6} discusses the consequences of some design-related decisions and proposed extensions.
\section{Related Work}
\label{sec:2}
A majority of existing literature casts the persistent coverage problem as an instantiation of the OP or the VRP.
An OP seeks to determine a path for an agent limited by a time- or distance-budget that visits a subset of all surveillance sites in an environment and maximizes the sum of associated scores collected.
An extension of this to multiple agents is known as the Team Orienteering Problem (TOP)~\cite{vansteenwegen2011orienteering}.
There are a number of works that apply solutions of the OP to surveillance with aerial vehicles over a small number of surveillance sites~\cite{keller2016thesis}.
The formulation by Leahy et al. comes closest to our work because of its capability of assigning time windows to each coverage-zone as temporal logic constraints~\cite{leahy2016persistent}.
%

Given a number of agents at a depot and distances among all surveillance sites and the depot, the VRP seeks to find a minimum-distance tour for each agent such that it visits each site at least once. Michael et al. address persistent surveillance with aerial vehicles as a VRP with modifications to model continuous site visits~\cite{golden2008vehicle, stump2011multi}.


The OP and VRP are both variants of the Traveling Salesman Problem (TSP) and thus NP-hard.
Moreover, in our case, formulating the problem as a TOP or VRP is not enough since our problem also has an element of 2D coverage.
One could consider every discrete part of the environment as a surveillance site in a TOP or VRP to enable 2D coverage, but this is impractical due to increasing computational complexity with the number of surveillance sites.

Smith et al. tackle a very similar problem by assuming pre-planned paths, decoupling path-planning and speed-control for deconfliction~\cite{smith2011sweeping}.
Planning paths for 2D coverage has been extensively studied in robotics~\cite{galceran2013survey}.
Typical solutions to static coverage involve environmental partitioning via Voroni tessellations and feedback control laws with local collision-avoidance schemes~\cite{schwager2009decentralized}.
Environmental partitioning is also a recurring strategy for dynamic and persistent coverage~\cite{MRC2kapoutsis2017darp}.
However, such partitioning restricts individual robots to specific parts of the environment.

For real-world settings, robot path-planners must adhere to kinodynamic motion-constraints.
Thakur et al. solve the TOP and path-planning simultaneously to generate 3D ($x,y,\theta$) space-parametrized trajectories, but without temporal deconfliction~\cite{thakur2013planning}. Many also formulate planning for persistent surveillance as a mixed-integer linear program (MILP)~\cite{bellingham2002coordination}.
However, such MILP formulations introduce limitations that are typically managed by planning for constant-speed paths~\cite{ademoye2006trajectory}.


A vital part of our framework is to ensure UAVs fly to appropriate goals, so that they simultaneously cover different parts of the environment.
We adapt frontier-based exploration to assign such goals to UAVs. Yamauchi et al. first proposed this for single and multiple robots based on an occupancy-grid representation of the environment~\cite{yamauchi1998frontier}.
Many extensions have been proposed since then, which combine the information gain or expected benefit of the goal and the distance to it, also called next-best-view approaches~\cite{adler2014autonomous}.
Zhu et al. also apply this to exploration and coverage with micro aerial vehicles (MAVs)~\cite{zhu2015mavs}.
We use frontier-based exploration to trade off between cell-proximity and cell-criticality (defined formally in Sec.~\ref{sec:3}). To be able to do this online, we use a search-based approach similar to Butzke et al.~\cite{butzke2011planning}.
%


\section{Problem Formulation}
\label{sec:3}
\begin{figure}[t]
  \centering
  \includegraphics[width=0.7\textwidth]{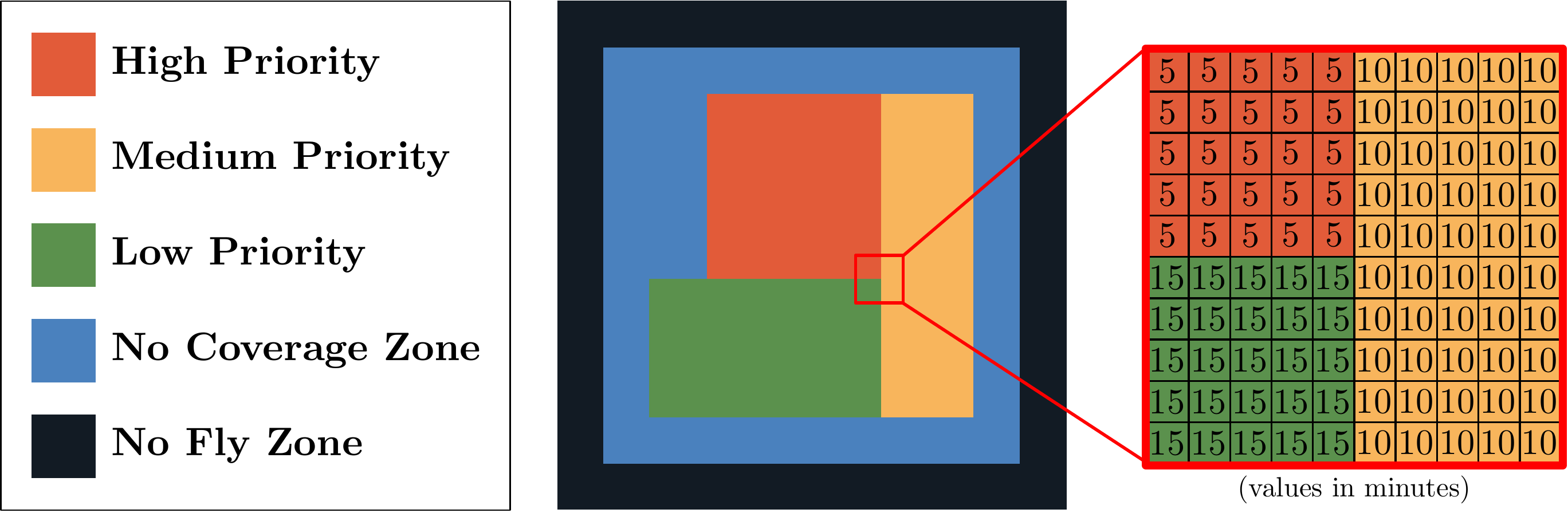}
\caption{\small{An example of a mission-map $\mathcal{M}$, where each cell is colored according to its lifetime. For example, a green cell with a lifetime of $15$ minutes implies that no more than 15 minutes should pass between two consecutive times a UAV covers it.}}
  \label{fig:priorities}
\end{figure}

Our problem definition is characterized by a mission-map $\mathcal{M}$, and a set of $N$ UAVs $\{U_1, \ldots, U_N\}$, tasked with covering the area represented by $\mathcal{M}$.
Each UAV $U_k$ is a kinodynamically constrained system, with an associated coverage- or sensor-radius~$r_k$.
Let the cell at row $i$ and column $j$ of $\mathcal{M}$ be $c_{i,j}$.
A UAV $U_k$ is at cell $c_{i,j}$ if the projection of its reference point onto the $x,y$ plane (denoted by $U_k^{\text{loc}}$) lies in cell $c_{i,j}$.
A cell is said to be covered by a UAV flying over it if any point on the cell is at at most an $r_k$ distance from $U_k^{\text{loc}}$.
This effectively assumes a circular sensor-footprint of radius $r_k$ centered around $U_k^{\text{loc}}$.
Each cell $c_{i,j}$ is associated with two temporal properties---its \emph{lifetime} $\ell(c_{i,j})$ and its \emph{age} $a(c_{i,j})$.
The age of a cell is the time passed since the cell was last covered by a UAV, while its lifetime is a desired bound on its age (as shown in Fig.~\ref{fig:priorities}).

There are two types of cells in our problem specification: (i)~standard cells (that UAVs need to cover), and (ii)~obstacle cells (that UAVs cannot fly over). Based on this, we define the following maps:
\begin{enumerate}
    \item Priority Map, $\Pmap = \left\{c_{i,j} : c_{i,j} \text{ is a standard cell} \wedge \ell(c_{i,j}) < \infty\right\}$. In other words, \Pmap is the set of all standard cells that a UAV will need to cover in finite time since the start of the mission.
    \item No-coverage Map, $\Nmap = \left\{c_{i,j} : c_{i,j} \text{ is a standard cell} \wedge \ell(c_{i,j}) = \infty\right\}$. In other words, \Nmap is the set of all standard cells in the mission-map that a UAV can fly over, but does not need to cover.
    \item Obstacle Map, $\Omap = \left\{c_{i,j} : c_{i,j} \text{ is an obstacle cell}\right\}$.
\end{enumerate}

The mission-map can then be defined as $\mathcal{M} = \Pmap \cup \Nmap \cup \Omap$.
Fig.~\ref{fig:priorities} contains an example of a mission-map with three different priority-levels.
Given a mission-map and a set of UAVs, our problem is to compute UAV-plans such that the following properties hold:
\begin{enumerate}[label={\textbf{P\arabic*}},leftmargin=1.0cm]
    \item \label{prop:feasibility}
       Feasibility---the motion of each UAV adheres to its kinodynamic constraints.
    \item \label{prop:deconfliction}
       Deconfliction---no two UAVs collide. In our specific setting, all UAVs are constrained to fly at the same altitude, so we say that two UAVs $U_p$ and $U_q$ collide if they come within some predefined distance \dmin of each other.
    \item \label{prop:persistence}
       Persistence---the age of each cell is smaller or equal to its lifetime.
    \item \label{prop:flexibility}
       Flexibility---UAVs can be dynamically added or removed to the system.
\end{enumerate}
\ref{prop:feasibility} and~\ref{prop:deconfliction} are hard constraints that must always be satisfied. \ref{prop:persistence} depends on the mission specification---the number of UAVs deployed, their capabilities (speed, sensor radii etc.) and the mission-map (size and lifetime of each cell). Our framework supports \ref{prop:flexibility} without affecting \ref{prop:feasibility}~and~\ref{prop:deconfliction}, however it will affect (help or hinder) our ability to satisfy \ref{prop:persistence}.

\section{Approach}
\label{sec:4}
%
Our approach, depicted in Fig.~\ref{fig:block_diagram}, consists of
a \emph{System Manager} (\SM) which serves as an interface between the centralized planning framework and individual UAVs.
It maintains the priority map~\Pmap by updating locations of individual UAVs\footnote{Our framework also allows for static obstacles and no-coverage zones to change during a mission, but this is beyond the scope of this paper.}.
%
Every planning cycle, the \SM passes the three maps
to the \emph{Prioritized Planner} (\PP).
%
The \PP then iterates over all the UAVs in a round-robin fashion.
For each UAV $U_k$, it generates a local copy of $\Pmap_k$ that accounts for the committed plans of the other UAVs.
This priority map $\Pmap_k$ is used by the \emph{Goal Assigner} (\GA) to assign a goal (in terms of $x,y$ location) for the UAV $U_k$.
A kinodynamically feasible path to reach this goal is then computed for $U_k$ using the \emph{Goal Planner} (\GP).

A key notion used within our framework is that of \emph{committed plans}, which are collision-free plans each UAV is committed to execute.
While the \GP might compute plans of long durations (if the goal is far away or plan execution requires long, feasible maneuvers), only a portion of this plan will be added to the committed plan, such that it is at most~\tmax long in time.
The value of~\tmax must be decided based on the mission since the following effects introduce a trade-off:
\begin{enumerate}
    \item Effects of a high value for \tmax:
    \begin{itemize}[leftmargin=*,noitemsep]
        \item \label{myopic}
        The longer the committed plans are, the less reactive UAVs are to any map updates including updates by an operator.
        %
        \item \label{doubly_cover}
        The path computed by the \GP might take $U_k$ through areas coinciding with parts of the committed plans of $U_{j\neq k}$ before $U_{j\neq k}$ covers them.
        This results in redundant coverage.
        The chances of this happening increase as \tmax increases.
    \end{itemize}
    \item Effects of a low value for \tmax:
    \begin{itemize}[leftmargin=*,noitemsep]
        \item \label{short_buffer}
        The shorter the committed plan, the lesser the chance of deviating from it during execution due to imperfect controllers.
        However, this increases the chances of UAV $U_k$ executing the committed plan before the next planning cycle ends, leaving the UAV's controller with no plan to execute.
        We handle this by ensuring that the UAV can execute a stopping-maneuver in this situation, but repeated execution of stopping-maneuvers is undesirable.
        \item \label{thrashing}
        Having short committed plans increases the frequency of goal assignment for a UAV.
        This causes two adverse effects:
        (i) A UAV may be assigned a new goal that is different from its previous one, which would cause it to abandon trying to reach the previous goal.
        (ii) Even if the same goal is assigned, it might lead to ``plan thrashing''.
        A UAV's complex dynamics may restrict short plans from letting it make significant progress towards a goal if complicated maneuvers are required.

        %
    \end{itemize}
\end{enumerate}

\subsection{System Manager}\label{subsec:SM}
The System Manager (\SM) is the communication interface between the centralized planner and the individual UAVs, shown in Fig.~\ref{fig:block_diagram} (left):
The \SM sends
the mission-map $\mathcal{M}$ to the \PP,
and it receives the updated committed plans from the \PP and sends them out to the respective UAVs.
It maintains the priority map~\Pmap by: (i)~resetting the age of any standard cells that have been covered since the last update to zero, and (ii)~incrementing the age of all other standard cells.

\begin{figure}[t]
    \centering
    \begin{subfigure}{.20\textwidth}
        \includegraphics[width=\textwidth]{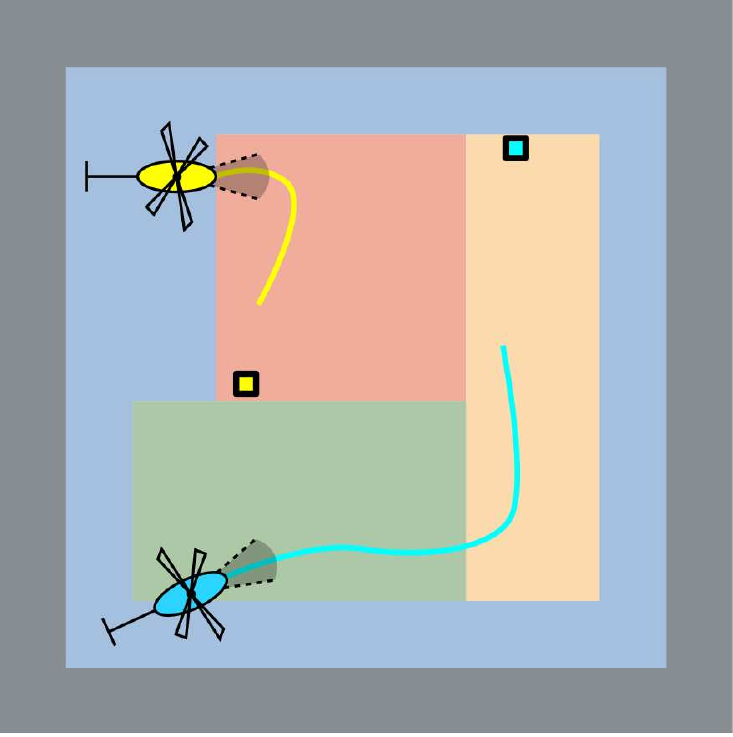}
        \caption{}
        \label{fig:pp_init}
    \end{subfigure}\hfill
    \begin{subfigure}{0.20\textwidth}
        \includegraphics[width=\textwidth]{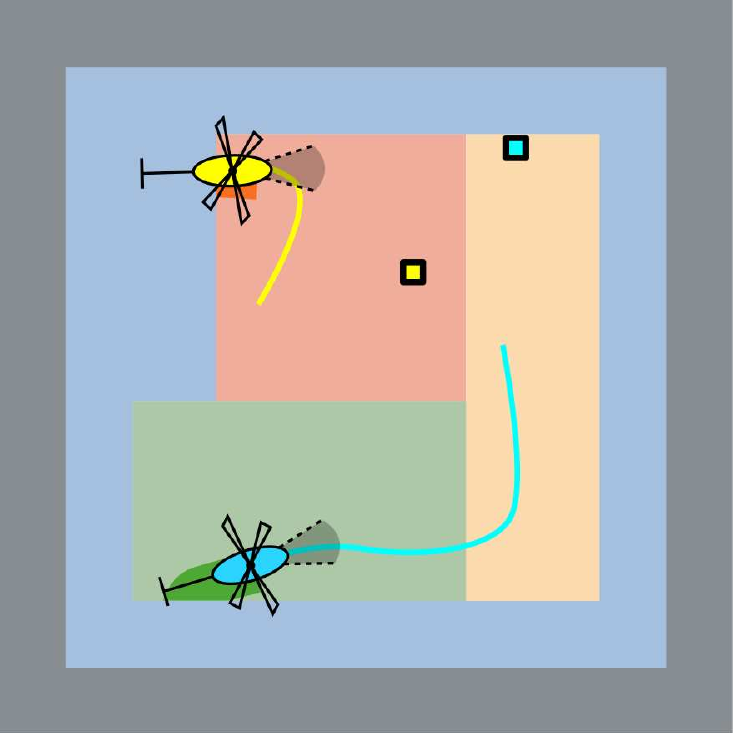}
        \caption{}
        \label{fig:pp_ga_yellow}
    \end{subfigure}\hfill
    \begin{subfigure}{0.20\textwidth}
        \includegraphics[width=\textwidth]{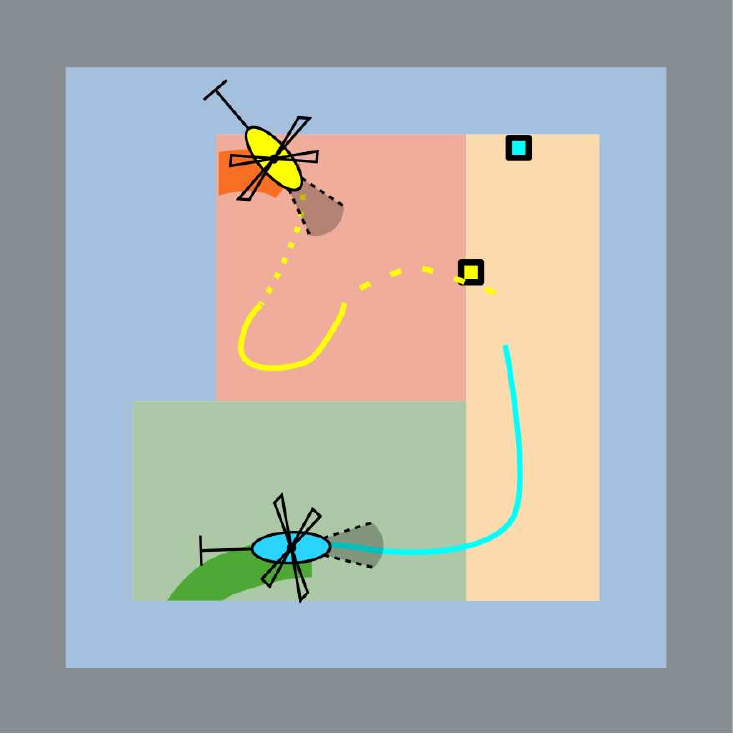}
        \caption{}
        \label{fig:pp_gp_yellow}
    \end{subfigure}\hfill
    \begin{subfigure}{0.20\textwidth}
        \includegraphics[width=\textwidth]{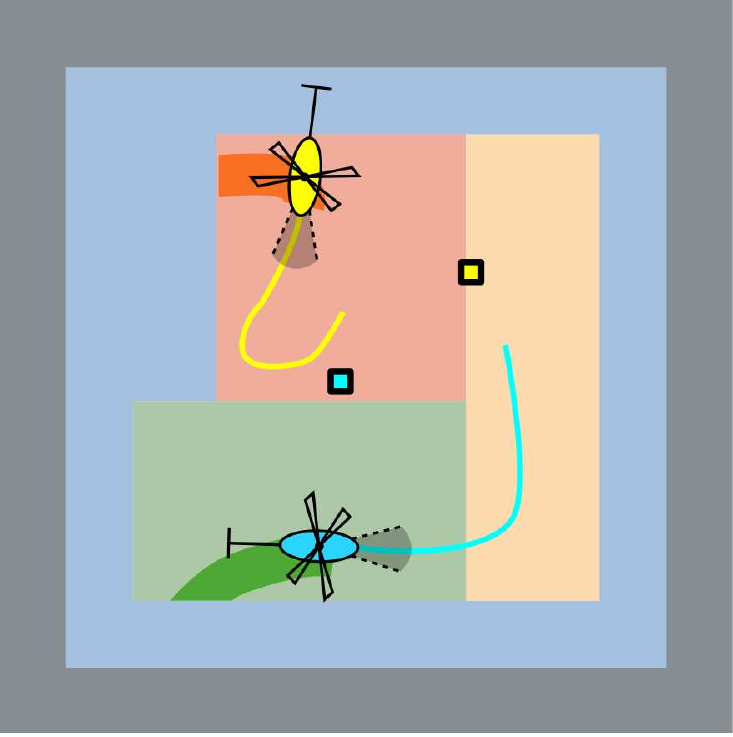}
        \caption{}
        \label{fig:pp_ga_blue}
    \end{subfigure}\hfill
    \begin{subfigure}{0.20\textwidth}
        \includegraphics[width=\textwidth]{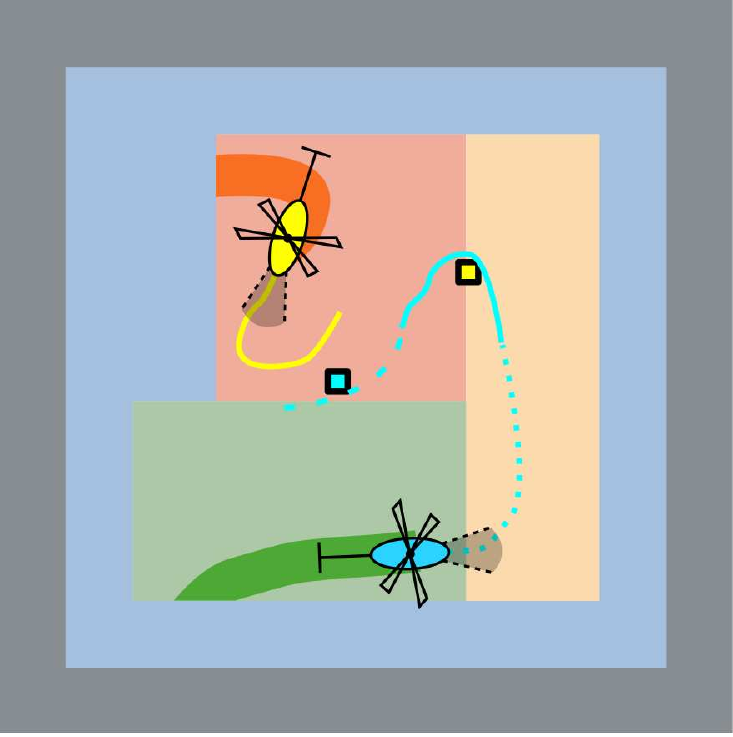}
        \caption{}
        \label{fig:pp_gp_blue}
    \end{subfigure}
    \vspace{-8pt}
    \caption{\small{Two successive executions of the \GA-\GP loop (the map $\mathcal{M}$ is colored with 50\% opacity).
             (\subref{fig:pp_init})~The solid yellow and blue lines show committed plans for both UAVs.
             (\subref{fig:pp_ga_yellow})~A new goal is assigned to the yellow UAV.
             (\subref{fig:pp_gp_yellow})~A path planned is for the yellow UAV (the old committed plan is a dotted line, the new committed plan is a solid line, and the discarded part of the new plan is a dashed line).
             (\subref{fig:pp_ga_blue})~The same as (\subref{fig:pp_ga_yellow}) but for the blue UAV.
             (\subref{fig:pp_gp_blue})~The same as (\subref{fig:pp_ga_yellow}) but for the blue UAV.}}
    \label{fig:buffers}
\end{figure}

\subsection{Prioritized Planner}\label{subsec:PP}
The Prioritized Planner (\PP) is a state-machine that controls our centralized planning framework by continuously executing \textproc{RunPlanner} from Alg.~\ref{alg:PP}.
During each execution of \textproc{RunPlanner}, which we call a \emph{planning cycle}, the \PP is responsible for the following tasks:
\begin{enumerate}[label={\textbf{T\arabic*}},leftmargin=1.0cm]
    \item \label{pp:update} Update its local copy of the committed plans and \Pmap to reflect the most recent information received from the \SM.
    \item \label{pp:uav} Determine which UAV will be planned for during this cycle.
    \item \label{pp:plan} Perform a new goal assginment for the UAV and plan a path to it\footnote{Task~\ref{pp:plan} is only done if the selected UAV's committed plan is shorter than~\tmax.}.
    \item \label{pp:commit} Communicate the result of \ref{pp:plan} throughout the framework.
\end{enumerate}
\ref{pp:update} is performed in Lines~\ref{alg:poll_sm}--\ref{alg:update_map} in Alg.~\ref{alg:PP}. The \PP clips the part of its local copy of committed plans that have been executed by the UAVs since the last planning cycle, and accounts for the part yet to be executed by resetting the age of all the cells in \Pmap that will be covered by the committed plans to zero.

%
%
We do not prioritize any one UAV over the others, which corresponds to the \PP selecting UAVs in a round-robin fashion, thus achieving \ref{pp:uav}.
While the \PP supports a prioritization over UAVs, in our experience, assigning equal priorities in this way showed satisfactory performance.

\ref{pp:plan} is achieved through one iteration of the \GA-\GP loop shown in Fig.~\ref{fig:pp_ga_gp}. For ease of understanding, we show two successive executions of this loop in Fig.~\ref{fig:buffers}. Fig.~\ref{fig:pp_ga_yellow} and~\ref{fig:pp_gp_yellow} correspond to Lines~\ref{alg:ga} and \ref{alg:gp} for the yellow UAV. Fig.~\ref{fig:pp_ga_blue} and~\ref{fig:pp_gp_blue} correspond to the same lines for the blue UAV.

\ref{pp:commit} is performed by Lines~\ref{alg:append}--\ref{alg:send}. First, we append to a UAV's committed plan to bring it up to \tmax in length. Second, \Pmap is updated to account for this new plan (same as in Line~\ref{alg:update_map}) by resetting the age of all the cells that will be covered. Finally, the newly committed plan is communicated to the UAV via the \SM.
\vspace{-1.3\baselineskip}
\begin{algorithm}[H]
\caption{\label{alg:PP} Prioritized Planner Loop}
\begin{small}
\begin{algorithmic}[1]
\Procedure{RunPlanner}{}
    \State  $N \gets$ number of UAVs \label{alg:init}
    \While{true}
        \State  Get latest $\mathcal{M} = \left\{\Pmap, \Nmap, \Omap\right\}$ and $U_{1:N}^{\text{loc}}$ from \SM \label{alg:poll_sm}
        \State  Update committed plans (local copies) by using $U_{1:N}^{\text{loc}}$ \Comment Clip \label{alg:clip}
        \State  Update \Pmap with committed plans \label{alg:update_map}
        \For{$k \gets 1:N$}
            \If{committed plan for $U_k$ is short}
                \State  $G_k \gets \textsc{GetGoal}\left(\Pmap\right)$  \Comment Goal assignment (\GA) \label{alg:ga}
                \State  $\pi_k \gets  \textsc{GetPlan}\left(\mathcal{M}, G_k\right)$ \Comment Goal planning (\GP) \label{alg:gp}
                \State  Update committed plan for $U_k$ with $\pi_k$ \Comment Append a portion of $\pi_k$ \label{alg:append}
                \State  Update \Pmap with committed plan for $U_k$ \label{alg:deconf}
                \State  Send committed plan for $U_k$ to \SM \label{alg:send}
            \EndIf
        \EndFor
    \EndWhile
\EndProcedure
\end{algorithmic}
\end{small}
\end{algorithm}
\vspace{-1.5\baselineskip}

\begin{figure}[t]
    \centering
    \includegraphics[width=.80\textwidth]{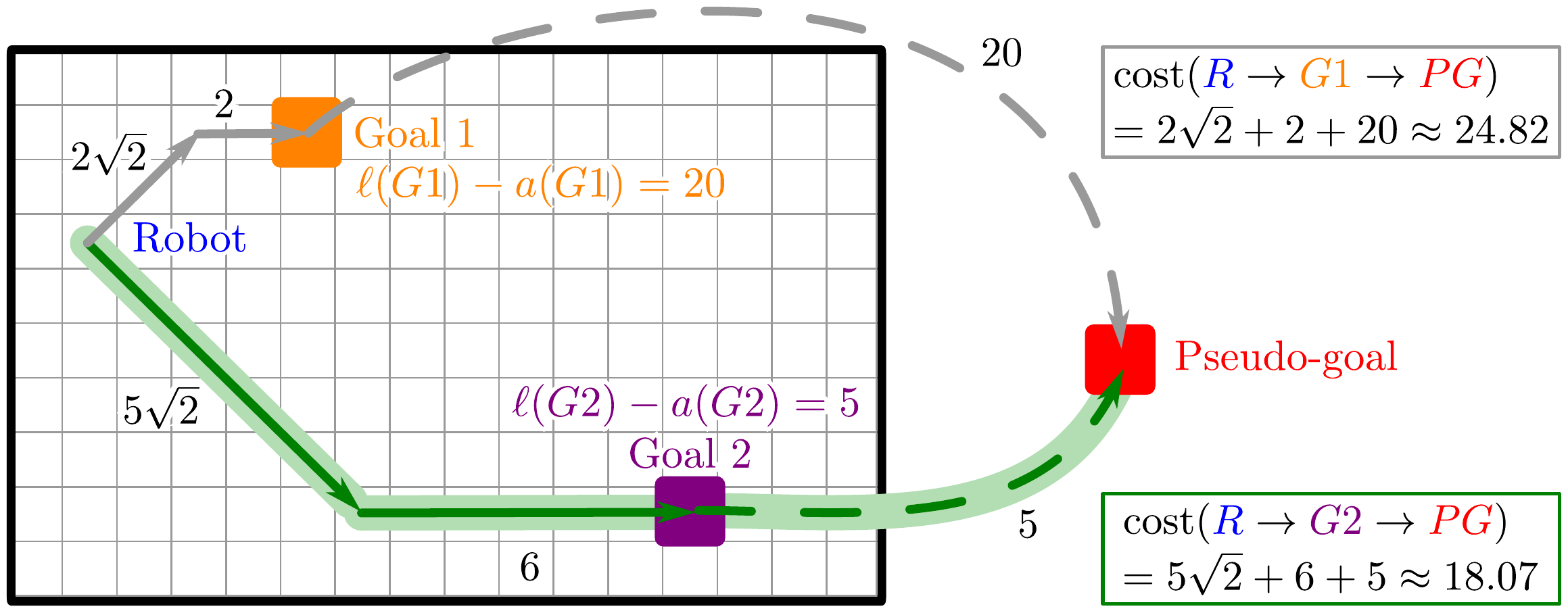}
    \caption{\small{Multi-goal Dijkstra search for goal assignment finds the least-cost path to a pseudo-goal connected to all potential goal locations for a robot. It considers the cost of reaching a potential goal from the robot's location, and the priority of a goal location reflected in the cost of the edge connecting it to the pseudogoal. In the example above, goal G2 is assigned to the robot.}}
    \label{fig:multi-goal}


\end{figure}
\subsection{Goal Assigner}\label{subsec:GA}

Recall that the Goal Assigner (\GA) selects a goal for each UAV to fly to.
The grid $\Pmap_k$ is the local copy of the Priority Map (\Pmap) that belongs to UAV $U_k$.
Each of the cells in the eight-connected grid that represents $\Pmap_k$ is a potential goal for $U_k$.
The \GA computes a goal-location~$G_\text{k}$ for each UAV, accounting for how urgent every cell in $\Pmap_k$ is and the location of the UAV $U_k$.
To simultaneously reason about the urgency of the goal and its distance from $U_k$, we build a graph by connecting a \emph{pseudo-goal} to every cell of the grid~$\Pmap_k$.
The pseudo-goal is an ``imaginary'' state connected by ``imaginary'' \emph{pseudo-edges} to all cells in the grid as shown in Fig.~\ref{fig:multi-goal}.
The cost of these pseudo-edges is proportional to $(\ell(i,j)-a(i,j))$\footnote{We lower-limit these costs by a constant to ensure they are strictly positive. Moreover, since the UAVs' sensors cover multiple cells, the cost of an edge between cell $c_{i,j}$ and the pseudo-goal is the average of $\ell(u,v)-a(u,v)$ for all cells $c_{u,v}$ covered when $U_k^{\text{loc}} = c_{i,j}$.}.
The costs of the rest of the edges between adjacent cells in the grid are equal to the Euclidean distance between them.

To find an optimal goal $G_k$, we first find an optimal path from $U_k^{\text{loc}}$ to the pseudo-goal on this graph using Dijkstra's search~\cite{dijkstra1959note}.
The optimal path minimizes the sum of the costs of the real-edges (which reflects the cost to reach the goal) and the cost of the pseudo-edge (which reflects the priority of the cell that the pseudo-edge connects).
The final goal $G_k$ for UAV $U_k$ is the parent of the pseudo-goal on this optimal path.

\subsection{Goal Planner}\label{subsec:GP}

Once a goal is assigned to $U_k$, the \GP computes a kinodynamically feasible path to it via search-based planning on a state-lattice~\cite{likhachev2009planning, pivtoraiko2005generating}.
The \emph{state-space} for each UAV includes its two-dimensional pose $(x, y, \theta)$, linear velocity $v$, and time $t$.
Assuming double-integrator dynamics for each UAV, we generate an action-space consisting of feasible motion primitives.
These primitives use cells of size $1 \si{\meter} \times 1 \si{\meter}$. This discretization is independent of the mission-map discretization.
We implicitly construct a graph using the action space during a weighted-A* search to the goal~\cite{hart1968formal, pohl1970heuristic}.
The search prunes away all transitions that correspond to trajectories that either intersect no-fly zones or collide with the committed plans of other vehicles in space or time.
We terminate the search as soon as a state is expanded whose incoming edge (from the predecessor on the found path) covers the goal cell (specifically, the trajectory corresponding to this edge contains a point whose distance to the goal cell is less than $r_k$).
We assign the time taken to execute an action as its edge-cost in the graph.
While our aim is to plan for minimum-time paths, it is also desirable for the UAVs to fly at high speeds whenever possible and avoid stopping.
For this reason, we incentivize actions with increasing velocities, penalize actions with decreasing velocities, and heavily penalize hovering.

\section{Experimental Setup and Results}
\label{sec:5}
%

We evaluate our framework in both simulation, where we assume perfect state estimation and control, and on real UAVs, which can deviate from their planned paths.
The UAVs can withstand winds of speeds up to \mPerSec{30}.
Thus, the effect of wind is negligible under normal conditions.
Accounting for large deviations from planned paths requires replanning and is part of future work.
Our framework uses an identical set of parameters in both cases.
We generate motion primitives with a maximum speed of \mPerSec{6} and a maximum turning rate of \degPerSec{6}.
We avoid adding angular velocity to the state-space by ensuring that these primitives start and end at zero angular velocity.
Two UAVs are deemed to have \textit{collided} if at any point in time the distance between these 2D locations is less than $d_\text{min} = 10\ \si{\meter}$.
The parameter $r_\text{k} = 15\ \si{\meter}$ defines the circular area directly underneath a UAV that is deemed \textit{covered} for any 2D location of the UAV.
We impose a planning timeout of $4\ \si{\second}$ on the Goal Planner, which is how long it is given to compute a plan.
We use an Euclidean-distance heuristic in the Weighted-A* search~\cite{pohl1970heuristic} with an inflation of $5$.
Fig.~\ref{fig:nagoyamap} and~\ref{fig:nagasakimap} shows the two coverage-maps used for the experiments presented in this paper.
%
%
%
\subsection{Evaluation Metrics}
We look at timing statistics for Goal Assignment ($t_{\text{GA}}$) and Goal Planning ($t_{\text{GP}}$),
number of state expansions, and number of expansions per second\footnote{These expansions refer to graph-node expansions in the Weighted-A* search.}. Since we desire continued movement from each UAV, we also look at the amount of time a UAV is stationary on average across the simulation run ($t_{\text{stopped}}$).
%
Additionally, we evaluate our framework's performance with respect to the \textit{criticality} of a cell, defined for any cell $(i,j)$ at timestamp $t$ as $C_{t}(i,j) = \frac{a_{t}(i,j)}{\ell(i,j)}$
, where $C_{t}: \mathbb{E}(2) \rightarrow \mathbb{R}$ is a time-varying measure of criticality.
$C_{t}(i,j)$ starts with a value of zero and as the cell \textit{ages} over time, $C_{t}(i,j)$ starts to increase. Once the age of a cell reaches its lifetime---namely, $C_{t}(i,j)$ equals one---we say that the cell has \emph{expired}.
%
%
If we average this value across all cells in the map to be covered, we obtain an estimate idea of how well a team of UAVs is covering the areas they have been assigned:
\vspace{-9pt}
\begin{equation}
    \bar{C}_t = \frac{1}{|\Pmap|} \sum_{(i,j) \in \Pmap} C_{t}(i,j)
    \vspace{-9pt}
\end{equation}\label{eq:avgcrit}
%
Given this measure of criticality, ideal behavior would be to keep the value of $\bar{C}_t$ constant over the course of a mission, a lower value being better.
\subsection{Simulation Experiments}
\label{sec:simexp}

\textbf{Hardware Platform} Simulation experiments were performed on a desktop computer running Ubuntu 16.04 and equipped with an Intel Core i7-4790K processor and 20 GB of RAM.

\begin{figure}[t]
    \centering
    \begin{subfigure}{0.49\textwidth}
        \includegraphics[width=\textwidth]{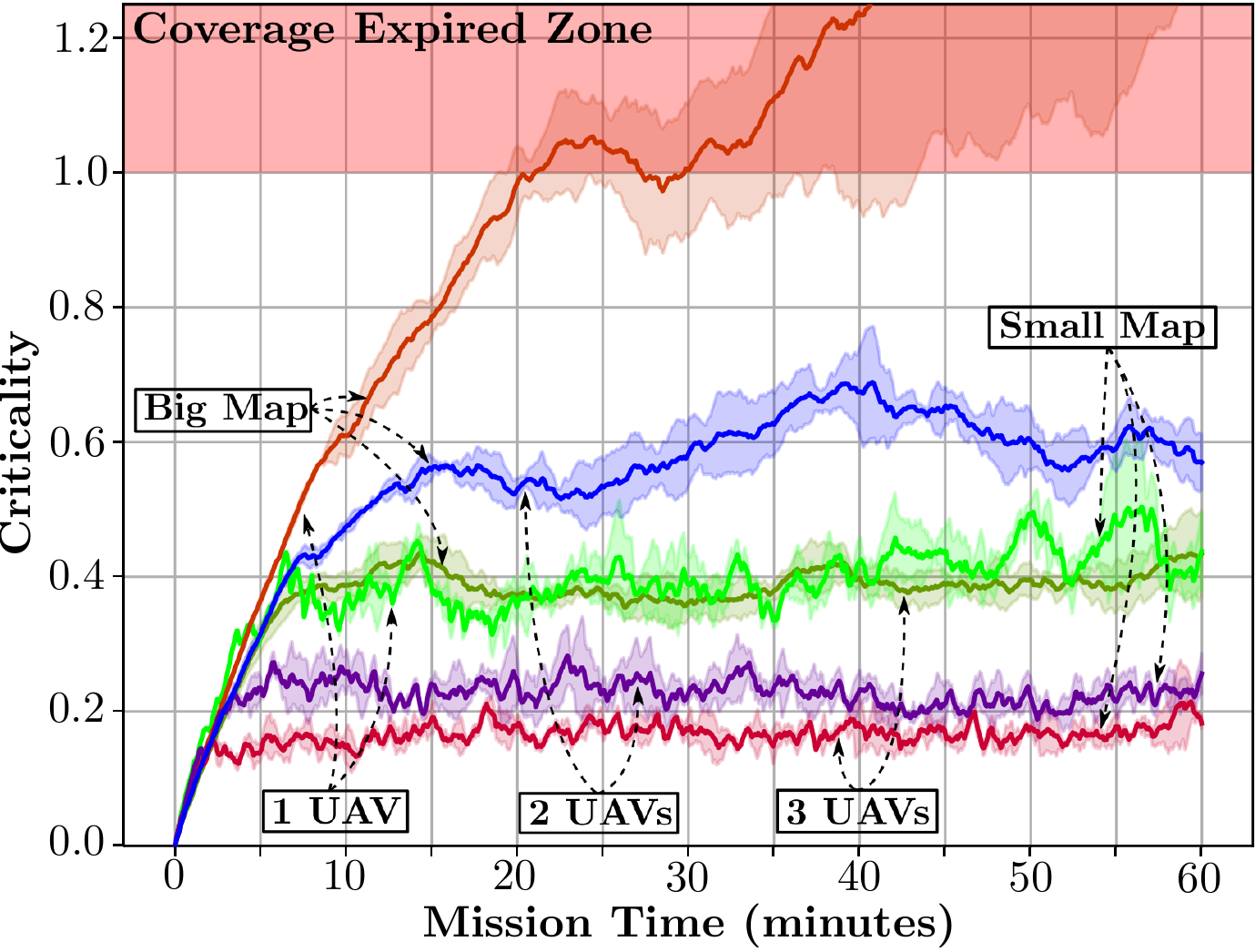}
        \label{fig:criticality}
    \end{subfigure}\hfill
    \begin{subfigure}{0.49\textwidth}
        \includegraphics[width=\textwidth]{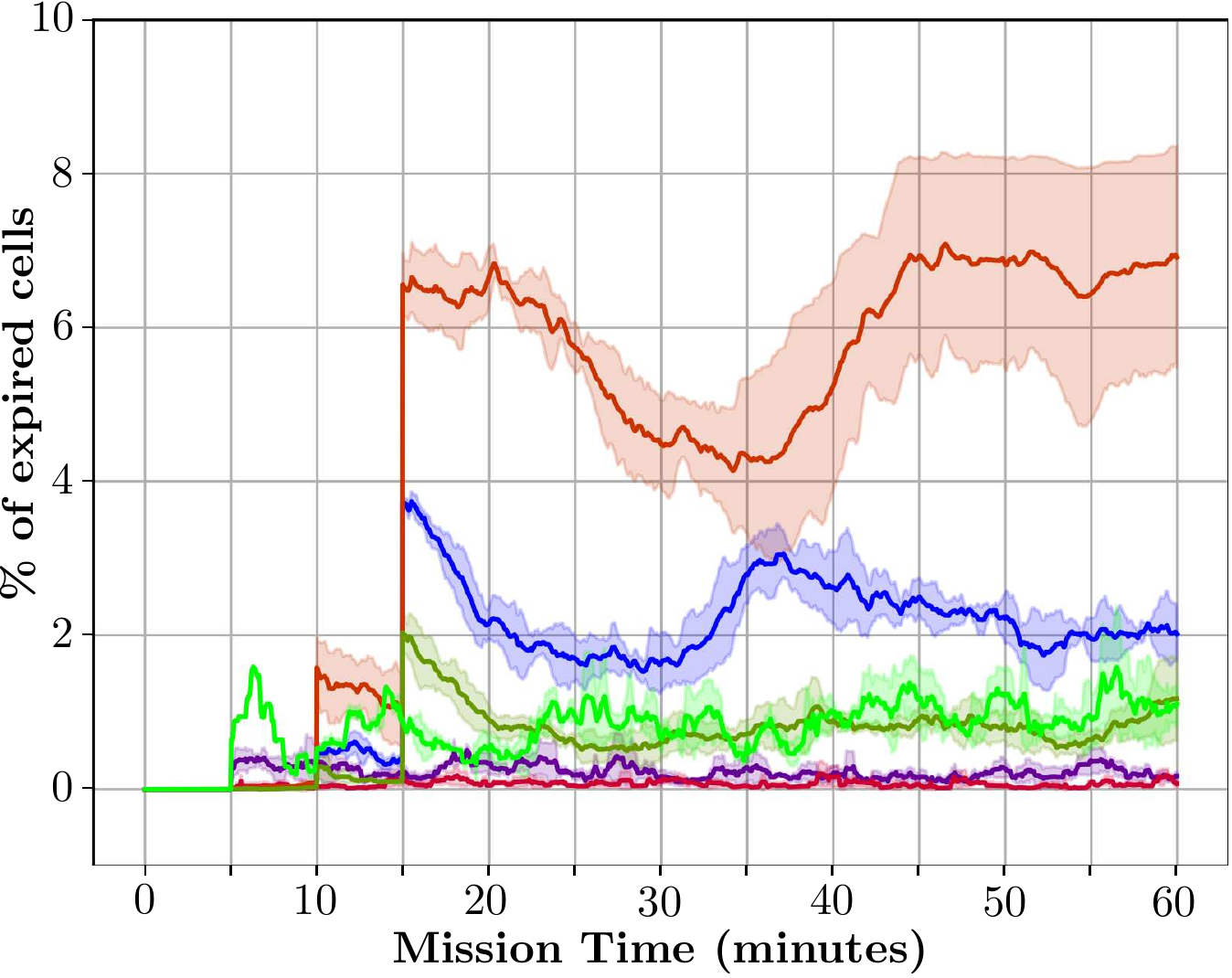}
        \label{fig:expired}
    \end{subfigure}
    \vspace{-19pt}
    \caption{\small{Left: Average criticality of cells in \Pmap during simulated experiments. Right: Spikes occur at 5, 10, 15 minutes since these are the three different cell lifetimes in the initial \Pmap. Colors are consistent across the two plots.}}
    \label{fig:coverage}
\end{figure}

\begin{figure}[t]
    \centering
    \begin{subfigure}{.16\textwidth}
        \includegraphics[width=\textwidth, keepaspectratio]{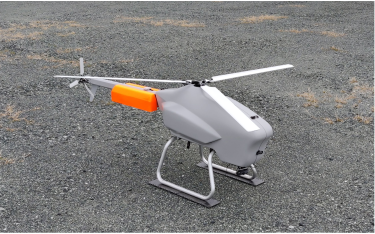}
        \caption{}
        \label{fig:real_uav}
    \end{subfigure}\hfill
    \begin{subfigure}{0.162\textwidth}
        \includegraphics[width=\textwidth, keepaspectratio]{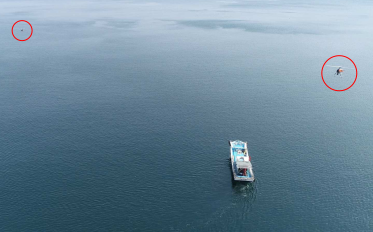}
        \caption{}
        \label{fig:inflight}
    \end{subfigure}\hfill
    \begin{subfigure}{0.20\textwidth}
        \includegraphics[width=\textwidth, keepaspectratio]{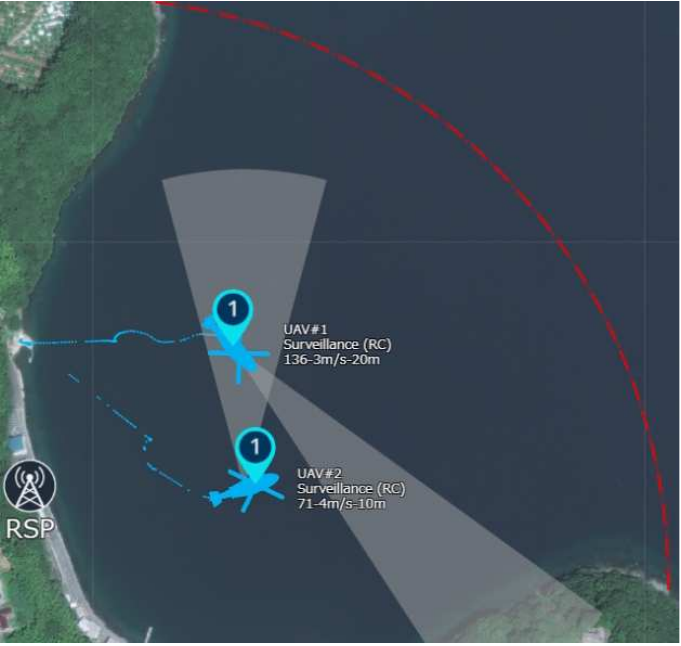}
        \caption{}
        \label{fig:hsi}
    \end{subfigure}\hfill
    \begin{subfigure}{0.19\textwidth}
        \includegraphics[width=\textwidth, keepaspectratio]{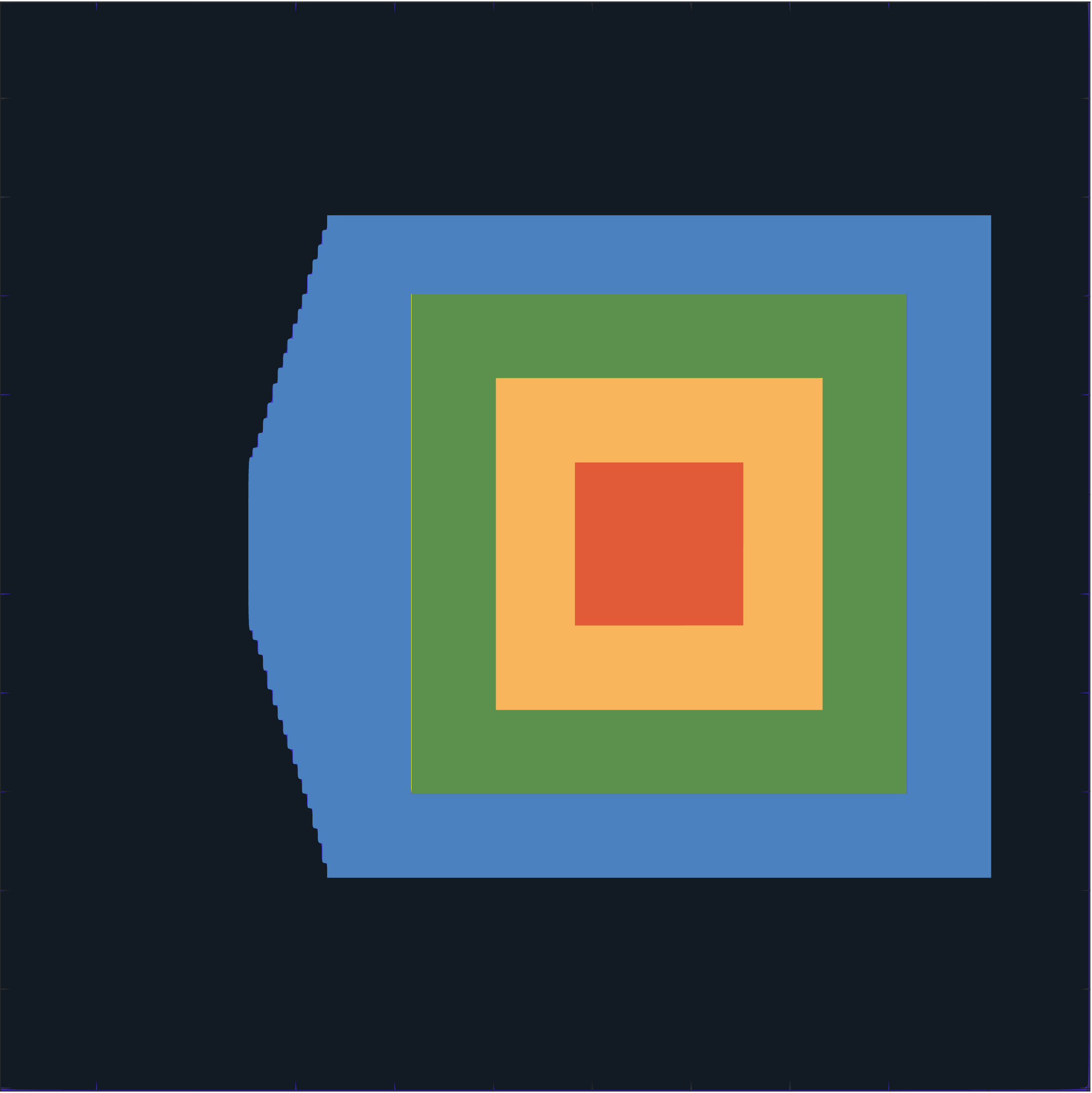}
        \caption{}
        \label{fig:nagoyamap}
    \end{subfigure}\hfill
    \begin{subfigure}{0.25\textwidth}
        \includegraphics[width=\textwidth, keepaspectratio]{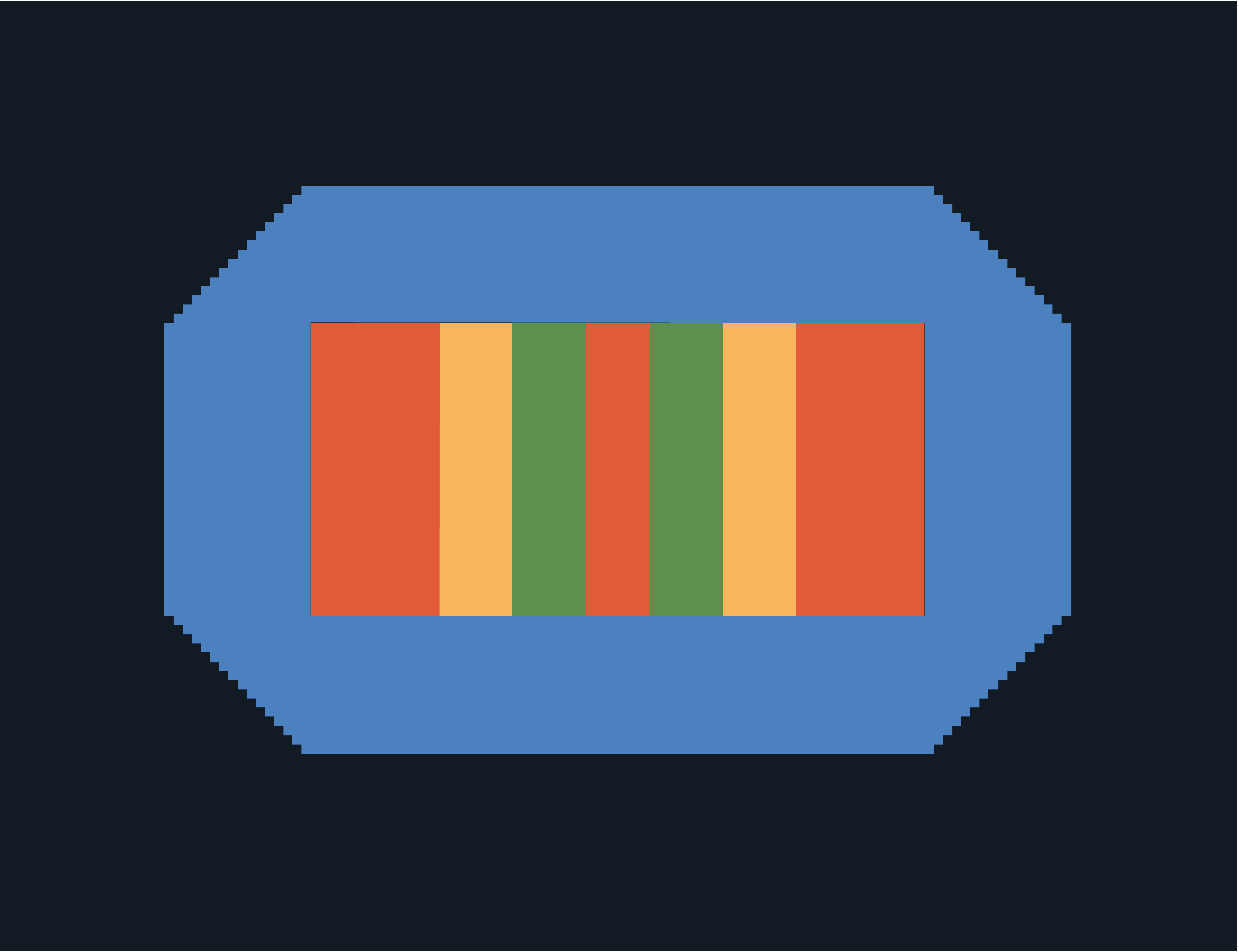}
        \caption{}
        \label{fig:nagasakimap}
    \end{subfigure}
    \caption{\small{(\subref{fig:real_uav}) UAV used in the experiments. (\subref{fig:inflight}) The two UAVs (circled in red) executing a mission. (\subref{fig:hsi}) GUI showing a satellite image of the mission site along with UAV data overlay.
    (\subref{fig:nagoyamap}) Map used for experiments (big; coverage zone roughly $400\text{m} \times 400\text{m}$).
    (\subref{fig:nagasakimap}) Map used for experiments (small; coverage zone roughly $200\text{m} \times 100\text{m}$)}}
    \label{fig:uavmission}
\end{figure}

\noindent\textbf{Planner Evaluation}
We present simulation results on both maps from Fig.~\ref{fig:uavmission} for $10$ minutes in Table~\ref{tab:simexp}.
We observe that by increasing the number of UAVs, Goal Assignment times are not affected, whereas Goal Planning times increase significantly since \GP must now compute deconflicting plans for a larger number of UAVs. Despite this, the total amount of stationary time spent by a UAV was at most $19.90\ \si{\second}$ in a $10$-minute period. Fig.~\ref{fig:coverage} shows a plot of $\bar{C}_{t}$ (Eq.~\eqref{eq:avgcrit}) over time during a 30-minute mission.
In all experiments, our framework is able to keep $\bar{C}_t$ below a value of one except where the number of UAVs was too small given the size of the map (experiment with one UAV and map from Fig.~\ref{fig:uavmission}).
This, along with the fact that all plots plateau over time, implies that cell-expiration is regulated on average.

\begin{table}\centering
\ra{1.3}
\begin{tabular}{crcrcccrccc}
\toprule
\textbf{Map} & \phantom{} & \multirowcell{3}{\textbf{Number}\\\textbf{of}\\\textbf{UAVs}} & \phantom{} & \multicolumn{3}{c}{\textbf{Timing} (ms)} & \phantom{} & \multicolumn{3}{c}{\textbf{Path Planning}} \\
\cmidrule{5-7} \cmidrule{9-11}
 &&  && \multicolumn{1}{p{1cm}}{\centering$t_{GA}$ \\} & \multicolumn{1}{p{1cm}}{\centering$t_{GP}$ \\} & \multicolumn{1}{p{1cm}}{\centering$t_{Total}$ \\} && \multicolumn{1}{p{2cm}}{\centering Number of \\ Expansions} & \multicolumn{1}{p{2cm}}{\centering Expansions \\ per second} & \multicolumn{1}{p{1cm}}{\centering $t_{\text{stopped}}$ \\ (\%)} \\ \midrule
%
Fig.~\ref{fig:nagoyamap} && 1 && 90 & 223 & 323 && 413 & 215 & 0                 \vspace{-3.2pt} \\
&& 2 && 90 & 796 & 898 && 187 & 226 & 2                     \vspace{-3.2pt} \\
&& 3 && 91 & 1326 & 1430 && 286 & 209 & 2                   \vspace{-3.2pt} \\ \midrule
Fig.~\ref{fig:nagasakimap} && 1 && 92 & 172 & 275 && 28 & 204 & 0                  \vspace{-3.2pt} \\
&& 2 && 91 & 1163 & 1265 && 200 & 188 & \textless \ 1       \vspace{-3.2pt} \\
&& 3 && 91 & 1433 & 1536 && 248 & 185 & 3 \\
%
\bottomrule
\end{tabular}
\caption{\small{Results of simulation experiments (values rounded down to the closest integer) and $t_{\text{stopped}}$ expressed as a percentage of total mission time.}}
\label{tab:simexp}
\end{table}

%
%
%

\begin{figure}[t]
    \centering
    \begin{subfigure}{.33\textwidth}
        \includegraphics[width=\textwidth]{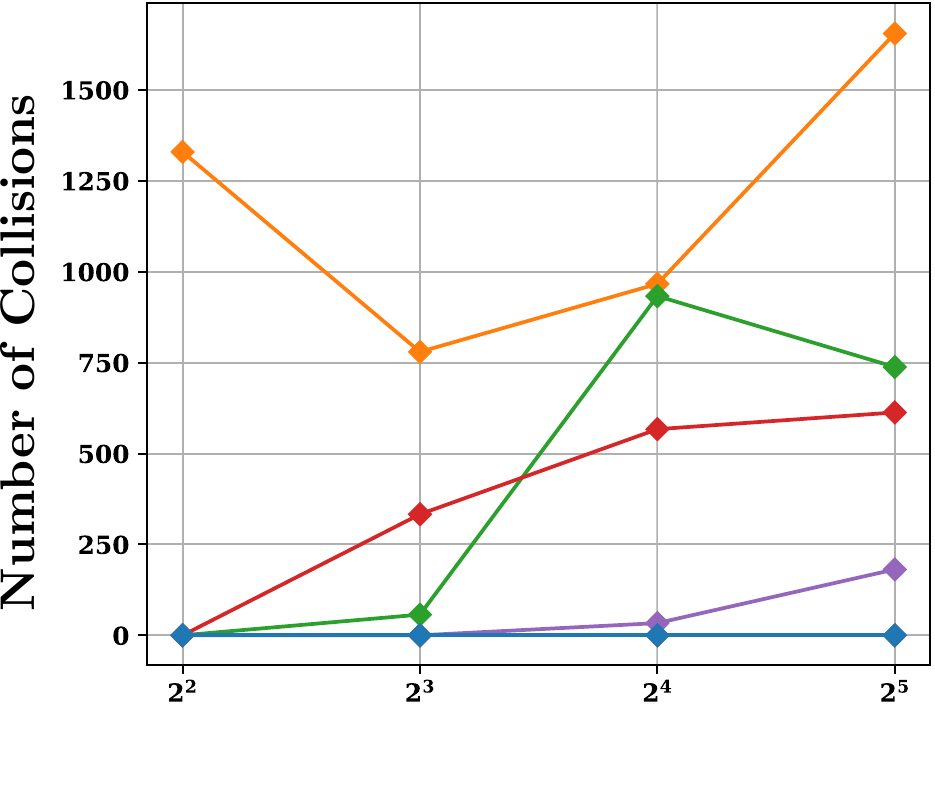}
        \caption{}
        \label{fig:buffer_length_collisions}
    \end{subfigure}\hfill
    \begin{subfigure}{0.33\textwidth}
        \includegraphics[width=\textwidth]{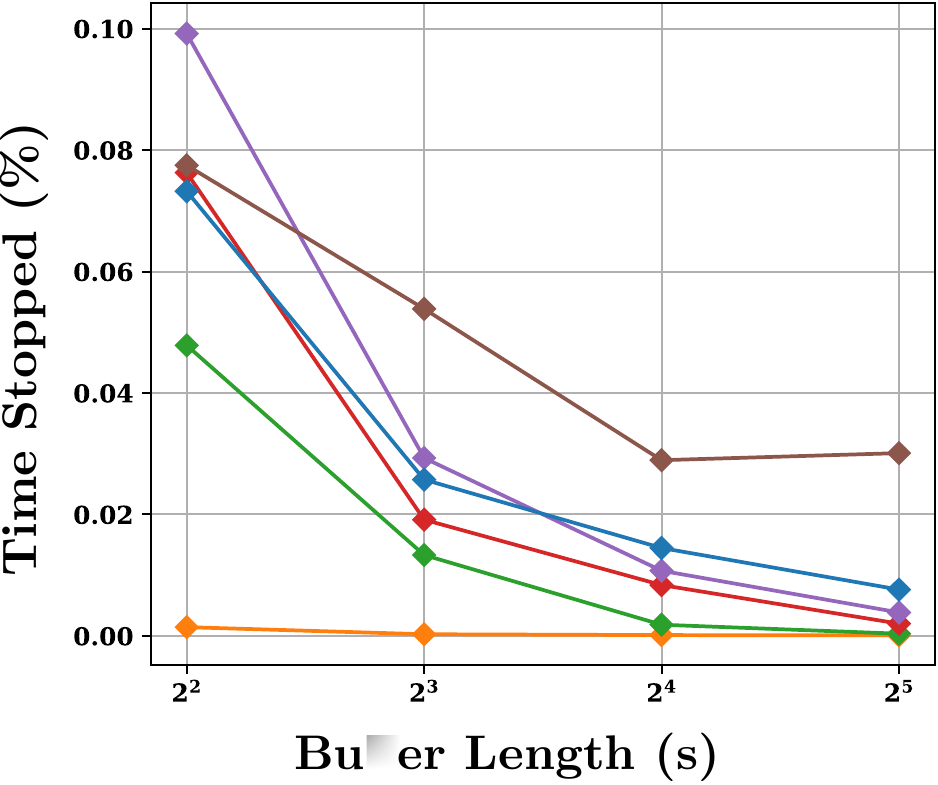}
        \caption{}
        \label{fig:buffer_length_stopped}
    \end{subfigure}\hfill
    \begin{subfigure}{0.33\textwidth}
        \includegraphics[width=\textwidth]{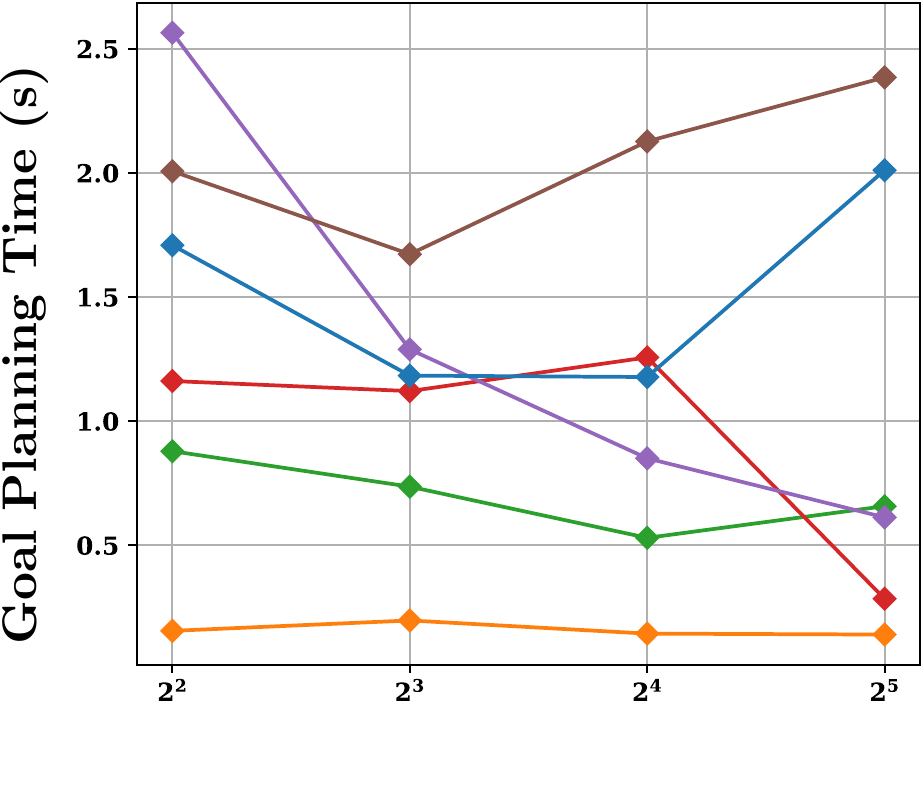}
        \caption{}
        \label{fig:buffer_length_gptime}
    \end{subfigure}
    \centering
    \begin{subfigure}{0.75\textwidth}
        \includegraphics[width=\textwidth]{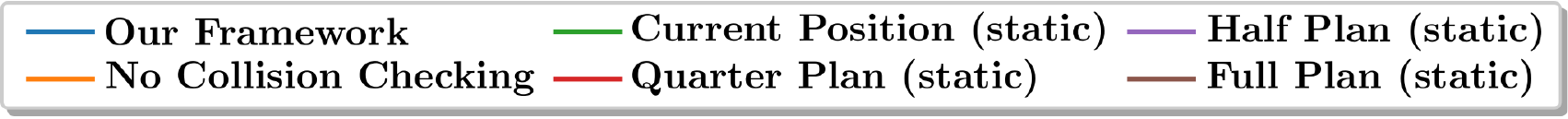}
        \label{fig:buffer_length_legend}
    \end{subfigure}
    \vspace{-15pt}
    \caption{\small{Effect of changing the committed plan length for different collision checking mechanisms. Since each curve represents a single mission execution, we do not show confidence intervals. The lines for \textit{Full Plan (static)} and \textit{Our Framework} coincide in (a).}}
    \label{fig:buffer_length}
\end{figure}

\noindent\textbf{Global Deconfliction}
%
%
Satisfactory global deconfliction relies on each UAV always having access to committed plans that are long enough. As discussed in Sec.~\ref{sec:4}, size of committed plans can significantly affect performance.
We empirically evaluate the need and effect of global deconfliction by varying the lengths of committed plans for different collision-checking schemes. These schemes are:
\begin{enumerate}
    \item Proposed Framework: Our approach with global deconfliction.
    \item No Collision Checking: The \GP performs  no collision-checking between UAVs.
    \item Current Position (static): The \GP considers the other UAVs' current positions as static obstacles.
    \item Quarter, Half and Full Plan (static): The \GP assumes the first quarter, half and whole of the committed plans of other UAVs to be static obstacles respectively.
\end{enumerate}
We plot the number of collisions between UAVs in Fig.~\ref{fig:buffer_length_collisions}, the average time a UAV is stationary in Fig.~\ref{fig:buffer_length_stopped}, and the time taken by the Goal Planner in Fig.~\ref{fig:buffer_length_gptime}, all against varying lengths of committed plans.
Fig.~\ref{fig:buffer_length} shows that our deconfliction scheme results in competitive planning times and also guarantees collision-free UAV movement with negligible stoppage.
Other collision-checking schemes are either overly conservative and compute convoluted, long-winding paths, or simply fail to avoid collisions.

\vspace{-4pt}
\subsection{Real-World Experiments}
\label{sec:realexp}

%





\textbf{Hardware Setup} For the real-world experiments, our planning framework was run on a Lenovo T470s laptop running Ubuntu 16.04 and equipped with an Intel Core i7-7600U processor and 20 GB of RAM.

\begin{figure}[t]
    \centering
    \begin{subfigure}{.51\textwidth}
        \includegraphics[width=\linewidth]{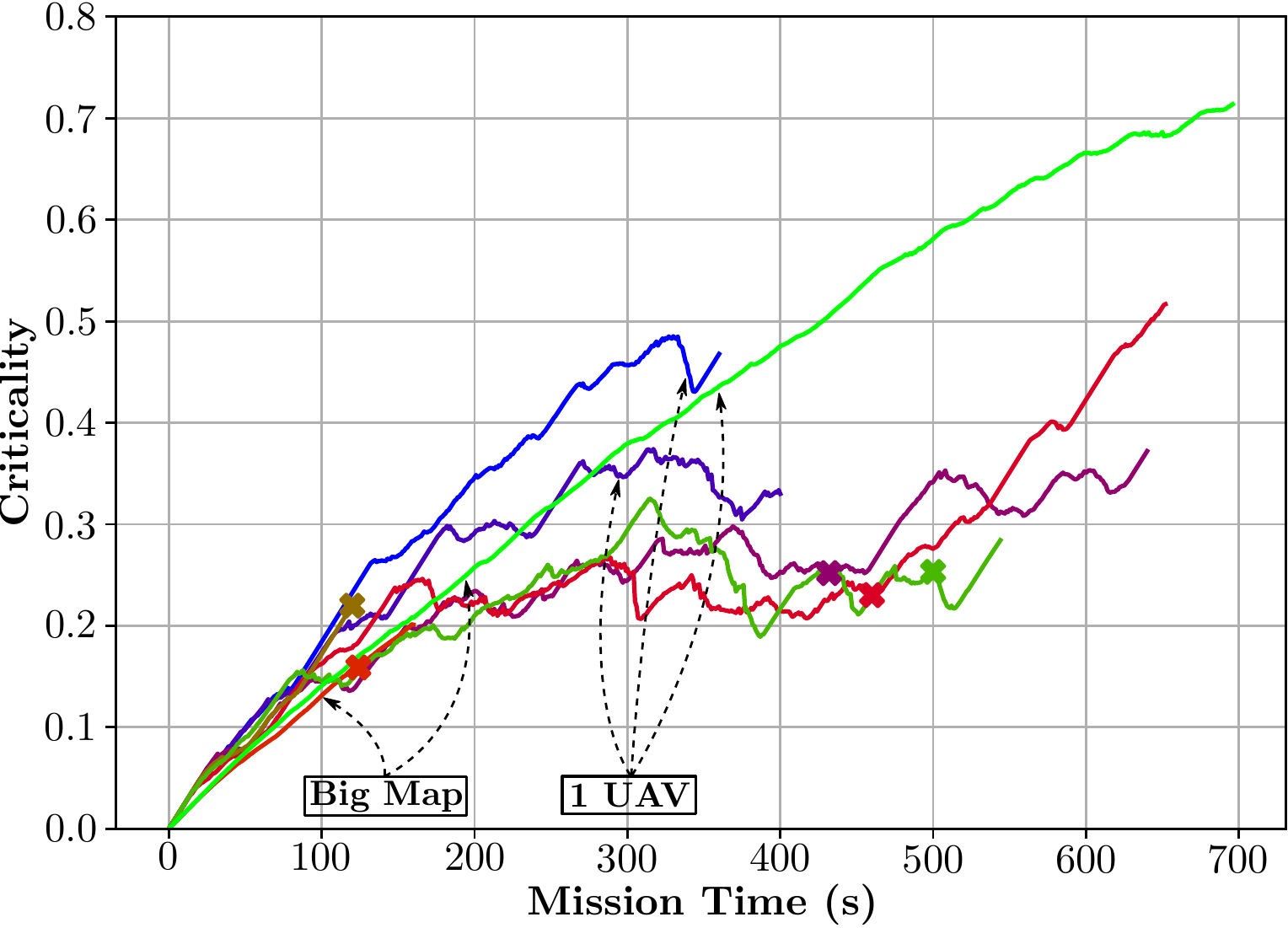}
        \caption{}
        \label{fig:coverage_comparison_real}
    \end{subfigure}\hfill
    \begin{subfigure}{.46\textwidth}
        \includegraphics[width=\linewidth]{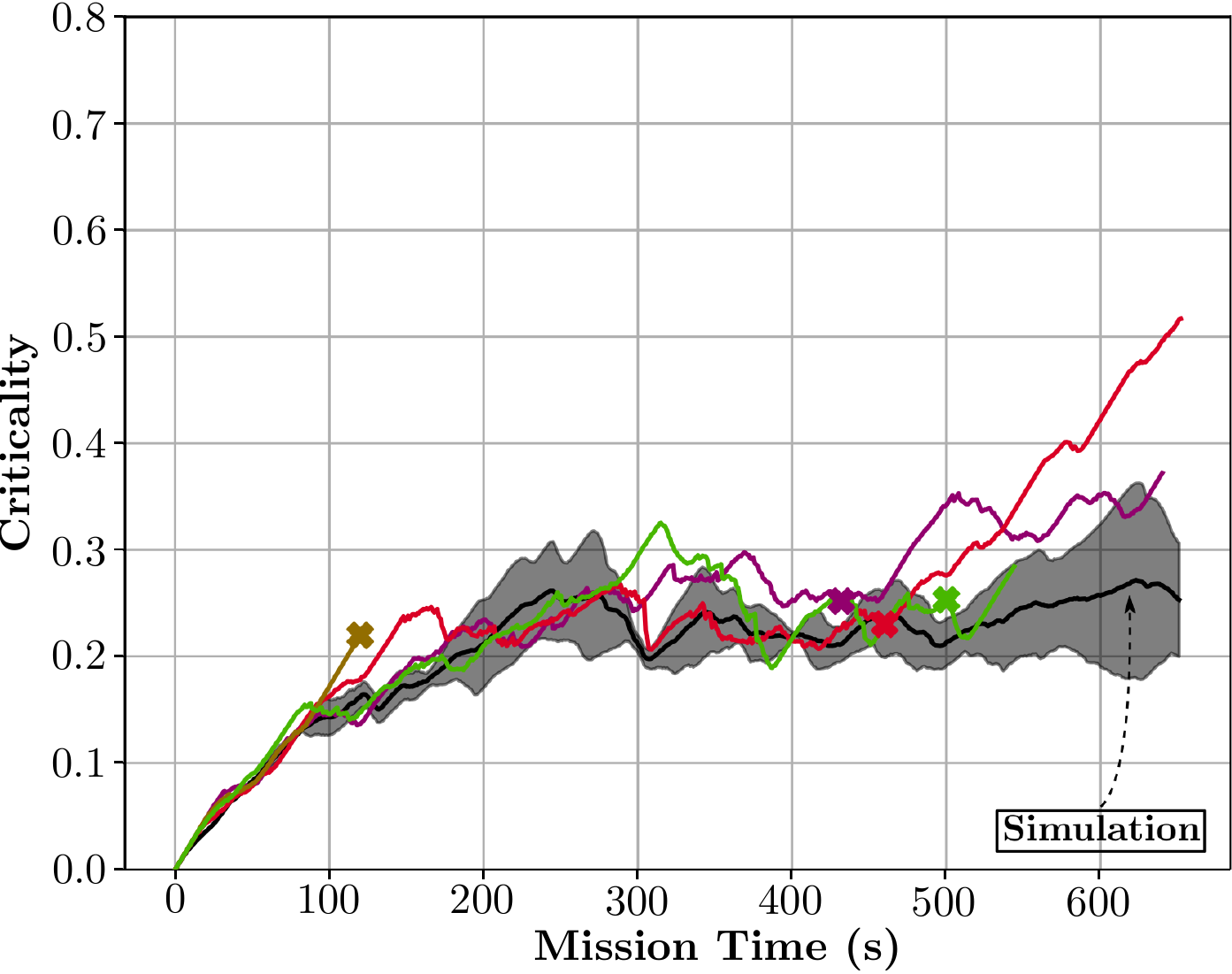}
        \caption{}
        \label{fig:coverage_comparison_withsim}
    \end{subfigure}
    \vspace{-10pt}
    \caption{\small{(a) Coverage criticality over time in the real-world: unless explicitly pointed to by arrows, the curves represent an experiment with two UAVs run on the smaller map. A cross ($\times$) on a curve indicates the point in time when one of the two UAVs was removed from the mission. (b) Comparison of all simulation and real-world missions executed in the small map from Fig.~\ref{fig:nagasakimap} by two UAVs. The black curve with confidence intervals corresponds to the simulated experiments.}}
    \label{fig:coverage_comparison}
\end{figure}

\textbf{Planner Evaluation} We present results from real-world experiments in Table~\ref{tab:realexp}.
Fig.~\ref{fig:coverage_comparison_real} plots $\bar{C}_t$ for eight real-world runs.
Additionally, dynamic removal of UAVs from the mission was tested in five of these runs, and the remaining UAVs were left undisturbed.
Beyond this point, we are only covering the area with one UAV.
Hence, it is expected to see criticality increase.
By iteratively planning and executing computed paths, we isolate our planner from the stochasticity of controller-based execution in the real world.
Our planner quantitatively performs just as well in the real-world as it does in simulation in spite of this stochasticity.
This is because our framework seeks latest information about UAV and map statuses from the real world and constantly uses it to repeatedly solve our myopic version of the full problem (as explained in Sec.~\ref{sec:4}).

%
\begin{table}\centering
\ra{1.3}
\begin{tabular}{crcrcccrccc}
\toprule
\textbf{Map} & \phantom{} & \multirowcell{3}{\textbf{Number}\\\textbf{of}\\\textbf{UAVs}} & \phantom{} & \multicolumn{3}{c}{\textbf{Timing} (ms)} & \phantom{} & \multicolumn{3}{c}{\textbf{Path Planning}} \\  
\cmidrule{5-7} \cmidrule{9-11}
 &&  && \multicolumn{1}{p{1cm}}{\centering$t_{\text{GA}}$} & \multicolumn{1}{p{1cm}}{\centering$t_{\text{GP}}$} & \multicolumn{1}{p{1cm}}{\centering$t_{\text{total}}$} && \multicolumn{1}{p{2cm}}{\centering Number of \\ Expansions} & \multicolumn{1}{p{2cm}}{\centering Expansions \\ per second} & \multicolumn{1}{p{1cm}}{\centering $t_{\text{stopped}}$ \\ (\%) } \\ \midrule
Fig.~\ref{fig:nagasakimap} && 1 && 105 & 240 & 363 && 37 & 186 & 0.00 \vspace{-3.2pt}\\
&& 2 && 117 & 1181 & 1325 && 162 & 152 & 7.12 \\
\bottomrule
\end{tabular}
\caption{\small{Results of real-world experiments averaged over 8 runs (rounded down to the nearest integer) and $t_{\text{stopped}}$ expressed as a percentage of total mission time (maximum over all runs).}}
\label{tab:realexp}
\end{table}
\vspace{-0.3cm}


\section{Conclusion and Future Work}
\label{sec:6}
We present and evaluate a planning framework for real-world, persistent coverage with multiple UAVs.
Our framework continuously decides where UAVs should fly and computes 
kinodynamically feasible, globally deconflicting 
plans.
We evaluate our framework in both simulated and real-world settings.
We also motivate and compare global deconfliction with weaker, more local collision-avoidance schemes.

In many practical settings like ours, state spaces are high-dimensional and time for deliberation is limited.
Planning times can be a bottleneck in these cases and cause delays.
While our stopping maneuvers handle such situations, a natural extension is to incorporate anytime planning~\cite{sun2008generalized}.

In the current framework, the goal assignment is based on priorities and is decoupled from goal planning.
This is greedy and not optimal. Better strategies to repeatedly cover previously observed coverage-zones can be learned from data and added as macro-actions in the planner.
\bibliographystyle{styles/spmpsci}
\bibliography{main}

\end{document}